\documentclass[12pt,onecolumn,letterpaper]{article}
\usepackage{cvpr}
\usepackage{times}
\usepackage{epsfig}
\usepackage{graphicx}
\usepackage{algpseudocode}
\usepackage{algorithm}
\usepackage{amsmath}
\usepackage{amssymb}
\usepackage{enumerate}
\usepackage{float}
\usepackage{subcaption}
\usepackage{xcolor}
\usepackage{multicol}
\usepackage[breaklinks=true,bookmarks=false]{hyperref}

\cvprfinalcopy % *** Uncomment this line for the final submission

\def\cvprPaperID{6958} % *** Enter the CVPR Paper ID here

\ifcvprfinal\fi

\title{Adaptive Hierarchical Down-Sampling for Point Cloud Classification}

\author{Ehsan Nezhadarya,
Ehsan Taghavi, Ryan Razani, Bingbing Liu and Jun Luo\\
Noah's Ark Lab, Huawei Technologies Inc.\\
Canada, Toronto \\
{\tt\small {$\{$ehsan.nezhadarya, ehsan.taghavi, ryan.razani, liu.bingbing, jun.luo1$\}$}@huawei.com}
}

\ifcvprfinal\fi
\begin{document}

\maketitle
%\thispagestyle{empty}

%%%%%%%%% ABSTRACT
\begin{abstract}
Deterministic down-sampling of an unordered point cloud in a deep neural network has not been
rigorously studied so far. Existing methods down-sample the points
regardless of their importance for the network output and often address down-sampling the raw point cloud before processing. As a result, some important points in the point cloud may be removed, while less valuable points may be passed to next layers. In contrast,
the proposed adaptive down-sampling method samples the points by taking into account the importance of each point, which varies according to application, task and training data. In this paper, we propose a novel deterministic, adaptive, permutation-invariant down-sampling layer, called Critical Points Layer (CPL), which learns to reduce the number of points in an unordered point cloud while retaining the important (critical) ones. Unlike most graph-based point cloud down-sampling methods that use $k$-NN to find the neighboring points, CPL is a global down-sampling method, rendering it  computationally very efficient. The proposed layer can be used along with a graph-based point cloud convolution layer to form a convolutional neural network, dubbed CP-Net in this paper. We introduce a CP-Net for $3$D object classification that achieves high accuracy for the ModelNet$40$ dataset among point cloud-based methods, which validates the effectiveness of the CPL.
\end{abstract}

%\begin{abstract}
%While several convolution-like operators have recently been proposed for extracting features out of point clouds, down-sampling an unordered point cloud in a deep neural network has not been rigorously studied. Existing methods down-sample the points regardless of their importance for the output. As a result, some important points in the point cloud may be removed, while less valuable points may be passed to the next layers. In contrast, adaptive down-sampling methods sample the points by taking into account the importance of each point, which varies based on the application, task and training data. In this paper, we propose a permutation-invariant learning-based adaptive down-sampling layer, called Critical Points Layer (CPL), which reduces the number of points in an unordered point cloud while retaining the important points. Unlike most graph-based point cloud down-sampling methods that use $k$-NN search algorithm to find the neighbouring points, CPL is a global down-sampling method, rendering it computationally very efficient. The proposed layer can be used along with any graph-based point cloud convolution layer to form a convolutional neural network, dubbed CP-Net in this paper. We introduce a CP-Net for $3$D object classification that achieves the best accuracy for the ModelNet$40$ dataset among point cloud-based methods, which validates the effectiveness of the CPL. 
%\end{abstract}
%%%%%%%%% BODY TEXT
\section{Introduction}\label{intro}
In most robotic applications, laser point cloud data plays a key role in perception of the surrounding environment. Autonomous mobile robotic systems in particular use point cloud data to train deep models to solve different problems, such as dynamic object detection, Simultaneous Localization and Mapping (SLAM), path planning, etc. Extracting features from  unordered point cloud using deep networks has become a highly active research field with the introduction of many new methods, such as PointNet \cite{qi2017pointnet}, PointNet++ \cite{qi2017pointnet++}, DGCNN \cite{wang2018dynamic}, PointCNN \cite{li2018pointcnn}, and SO-Net \cite{li2018so}. These methods are shown to be quite successful in point cloud classification benchmarks, such as ModelNet$40$ \cite{wu20153d}.

%In practical scenarios, the number of points in the point cloud associated with an object may be large, especially with high density sensors such as Velodyne-$64$ \cite{moosmann2011velodyne}. A car zipping by the mobile robot, for example, may correspond to thousands of LiDAR points. Methods such as PointNet \cite{qi2017pointnet} and DGCNN \cite{wang2018dynamic} pass the same number of input points into deeper layers to extract complex features. Without reduction of input size, however, much redundant information must be processed by the deeper layers, resulting in extra delay in the classification pipeline. \footnote{A key constraint here is that the unordered point cloud is processed in a way such that the order of points has no effect. This differs from the image domain where pixels are densely ordered in space, which naturally leads to acceleration through GPU data parallelism. Due to order-agnosticism and data sparsity, the recent methods for classification of unordered point cloud can not take full advantage of GPU parallelism.}

In practical scenarios, the number of points in the point cloud associated with an object may be quite large, especially as a result of using high density sensors such as Velodyne-$64$ \cite{moosmann2011velodyne}. One possible way to reduce computation is to down-sample the points in the point cloud as it gets passed through the network. A class of methods are proposed in which $k$-NN search \cite{peterson2009k} is used to find the neighbourhood for each point and down-sample according to these neighbourhoods. Such methods, however, trade one kind of expensive computation (neighbourhood search) for another one (processing large point cloud). 

In another family of works such as \cite{qi2017pointnet} and \cite{qi2017pointnet++}, the $3$D point cloud is processed directly, while other works transform point cloud to regular voxels such as methods in \cite{wu20153d, maturana2015voxnet, gadelha2018multiresolution}.  Transforming to regular voxels however, leads to loss of goemetric information and high computational complexity. Recently, the method introduced in RS-CNN \cite{liu2019relation} attempts to learn irregular CNN-like filters to capture local features for point cloud which achieves state-of-the-art accuracy in classification. In addition to the aforementioned papers, \cite{dovrat2019learning}, a deep learning network for point cloud sampling is presented. Method in \cite{dovrat2019learning} produces a down-sampled point cloud from raw and unordered input point cloud which is not guaranteed to be a subset of the original one. Therefore, a post-processing matching step is required, leading to a more complex system. 

In order to fully leverage a down-sampling method, what is highly needed is a deterministic content-sensitive but fast way of down-sampling an unordered point cloud that can be integrated into a deep neural network -- a technique similarly effective and efficient as max pooling in conventional CNN. In this paper, we introduce the \textit{Critical Points Layer} (CPL), which meets these requirements.

Unlike previous down-sampling methods that generate a set of points different from the input points, CPL not only \textit{selects} the output points from the input, but also down-samples the points within the network in a way that the \textit{critical} ones are not lost in this process. Unlike random sampling type of layers which generate randomly different set of points at inference time, CPL is a \textit{deterministic} layer, producing the same set of points after each run. It is \textit{invariant} to permutation of input points, i.e. order-agnostic. It is \textit{adaptive} in that it learns to down-sample the points during training and it is a \textit{global} method not limited to neighbourhood search, which makes it efficient.

%The CPL can be used along with any graph-based point cloud convolution layer \cite{learning2016} to form larger neural nets, dubbed CP-Nets. Our experiments show that CP-Nets can achieve state-of-the-art classification accuracy with computational cost kept at a reasonable level.

\section{Related Work}
% Talk about related graph down-sampling methods 

\subsection{Deep Learning on Point Clouds}
To cope with sparsity of point cloud, deep learning methods tend to voxelize space and then apply $3$D CNNs to the voxels \cite{maturana2015voxnet, zhirong15cvpr}. A problem with this approach is that network size and computational complexity grow quickly with spatial resolution. On the flip side, lower spatial resolution means larger voxels and higher quantization error. One way to alleviate this problem is to use octree \cite{wang2017cnn} or kd-tree \cite{klokov2017escape} instead of voxels. In \cite{klokov2017escape}, for example, a kd-tree of point cloud is built and then traversed by a hierarchical feature extractor, exploiting invariance of the point cloud at different spatial scale. However, such methods still rely on subdividing a bounding volume and fail to exploit local geometric structure of the points themselves. In contrast, point-based neural networks does not require converting point clouds to another format. Resolution loss is thus avoided \cite{yang2018foldingnet}. 
%In \cite{ravanbakhsh2016deep}, it is shown that given a symmetric function, such as max or summation, a permutation-invariant NN layer can consume unordered point cloud for object classification and semantic segmentation. PointNet \cite{qi2017pointnet} is a pioneering method of this kind. Because PointNet treats points independently and uses global aggregation, however, it still fails to fully take advantage of either local structures (such as corners) or geometric relations among neighbouring points. To address this limitation, PointNet++ \cite{qi2017pointnet++} partitions the points into small overlapping clusters before applying PointNet to each cluster independently. This method does exploit local structures captured by clustering, but still ignores within-cluster geometric relations.

\subsection{CNNs on Point Cloud as Graphs}
A promising way to exploit local geometric information is to model point cloud as graph of unordered (non-euclidean) points and then apply CNNs to it. This important research direction \cite{simonovsky2017dynamic} has two main variations.

\textit{Spectral methods} redefine spatial graph convolution as a multiplication in the spectral domain \cite{simonovsky2017dynamic, shuman2012windowed}. The first proposed methods along this line lack spatial locality of filters. Parameterization of convolution filters as Chebyshev polynomials of eigenvalues and approximately evaluating them lead to a computationally efficient way to create localized spatial filters \cite{defferrard2016convolutional}. Unfortunately, these filters are learnt in the spectral domain \cite{anderson1985eigenvalues} and thus have to be the same for all graphs in the dataset. This means that where graph structure varies in the dataset, such as point clouds, a graph learnt on one shape cannot generalize to others.

\textit{Local spatial filtering} \cite{simonovsky2017dynamic,niepert2016learning, levie2017cayleynets, velickovic2017graph, zhang2018end, boscaini2016learning,masci2015geodesic,monti2017geometric}, in contrast, employs spatial filters. A notion of local patch on graphs is employed to allow an operator similar to convolution to be applied to each local patch. Depending on the specific correspondence between filter weights and nodes in each local patch, we have variants such as MoNet \cite{monti2017geometric}, GCNN \cite{kipf2016semi} and DCNN \cite{atwood15}. Although much work has been done to apply spatial filtering for deep learning on general graphs, only a few methods, such as KCNet \cite{shen2018mining}, FoldingNet  \cite{yang2018foldingnet}, ECC \cite{simonovsky2017dynamic} and DGCNN  \cite{wang2018dynamic} use deep learning on \textit{point cloud} graphs.

\subsection{Point Cloud Down-Sampling in Deep Networks}
While graph convolution on point clouds has recently received great attention, down-sampling of the input point cloud remains largely unexplored. Such down-sampling is highly desirable for a few reasons: 
\begin{itemize}
\item Most graph convolution methods on point cloud use $k$-NN search to find the neighbourhood of each point. Thus, down-sampling the points cuts the computational cost for subsequent convolution layers.
\item Reducing the number of points in the network results in lower runtime memory usage.
\item Down-sampling can boost robustness to certain perturbations in the input data.
\end{itemize}
In typical point-based neural networks, such as PointNet and DGCNN, the number of points in the point cloud is fixed throughout the network. PointNet++ does down-sample the point cloud using farthest point sampling (FPS). However, because it generates overlapping partitions by finding the $k$-NN points around each sample point, much more computation is needed than PointNet \cite{qi2017pointnet++}.

KCNet \cite{shen2018mining} and FoldingNet \cite{yang2018foldingnet} down-sample the graph using a graph-based max-pooling that takes maximum features over the neighbourhood of each node using a pre-built $k$-NN graph (KNNG). However, these methods provide no guarantee that the most important points, which we call \textit{critical points}, will be passed to downstream. A point with less relevant features may be selected or generated, while an important one may be removed or devalued.

Moreover, the down-sampling used in some of these networks are static, where the sampling is only based on spatial locations of points in the input point cloud, but not on their corresponding learnt features. On the other hand, methods that do use feature space distance between points, such as PointNet++ \cite{qi2017pointnet++}, are computationally prohibitive as explained earlier. In another prominent method,  \cite{dovrat2019learning} introduces a deep learning approach for optimized point cloud sampling in which a NN model is trained to generate a down-sampled point cloud from the original dataset.

Finally, all these methods \textit{generate} a set of new points, instead of \textit{selecting} a subset of input points. This makes it difficult to track the contribution of each input point on the output.

In this paper, we introduce a computationally efficient \textit{Critical Points Layer} (CPL), which down-samples the points \textit{adaptively} based on the learnt features. CPL \textit{globally} filters out unimportant points while keeping the important ones, according to a point's level of contribution to the global max-pooling (max-reduced feature vector). The CPL is computationally very efficient because it does not need local nearest neighbour search. Moreover, since the feature vectors obtained from a graph convolution layer already contain the local neighbourhood information of each point that is important, the CPL yields a smaller subset without losing relevant information. In the next two sections, we explain the CPL in detail and report our experimental results.  

\section{Proposed Solution}
In this section, we propose two new adaptive down-sampling methods that can be used in deep neural networks. The focus of the proposed methods is to introduce a systemic solution to down-sample the points (or the feature vectors associated to them) in an arbitrary neural network architecture. This differs with methods such as \cite{dovrat2019learning}, in which a novel method is proposed to down-sample a specific dataset. These two new layers, named \textit{Critical Points Layer} (CPL) and \textit{Weighted Critical Points Layer} (WCPL), can efficiently down-sample the features related to an unordered point cloud, while being permutation invariant. In this section, CPL, WCPL and a systemic approach to use these two layers in deep neural networks and more specifically classification neural networks will be explained in details.

\subsection{Critical Points Layer (CPL)}

%TBE: A method to order points: Compare with sorting the output of MLP layers based on summation of feature values

Lets assume the input to the CPL is an unordered point cloud with $n$ points, each represented as a feature vector $\textbf{x} \in \mathbb{R}^d$, where $\mathbb{R}$ is the set of real numbers and $d$ is the dimension of the feature vector. The goal of CPL is to generate a subset of input points, called \textit{Critical Points} (CP), with $m\leq n$ points, each represented as a feature vector $\textbf{y} \in \mathbb{R}^l$, where $l$ is the dimension of the new feature vector. The critical points of a point cloud are the points with maximum information that are needed to be preserved in a down-sampling (or pooling) process. These points may be changed based on the task and application. 

\begin{figure*}%[t]
\label{fig_cpl_wcpl}
\centering
\includegraphics[trim = 25px 200px 210px 0px, clip=true,width=4.5in]{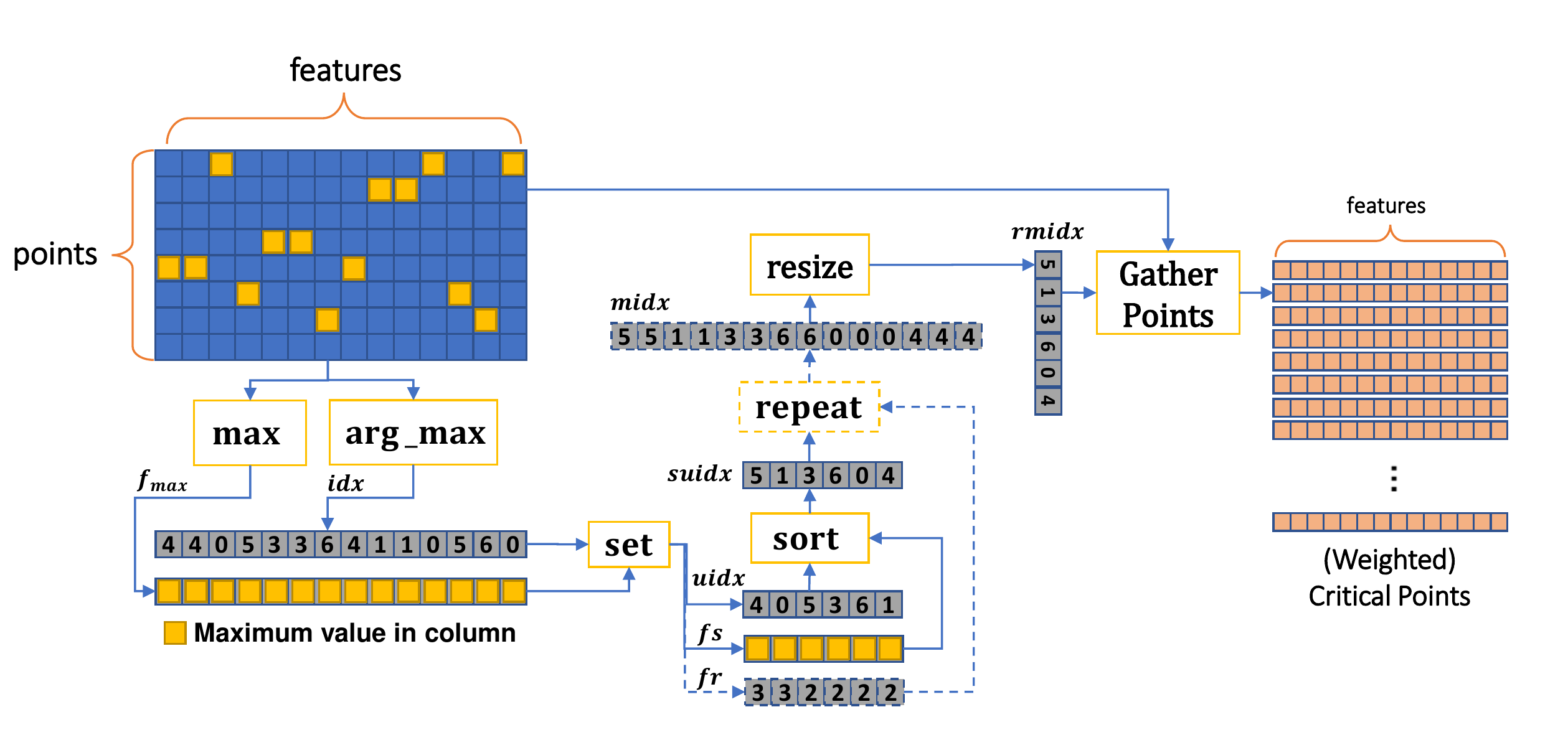}
\caption{Critical Points Layer (CPL) [solid lines] and Weighted Critical Points Layer (WCPL) [Solid and dashed lines]}
\label{fig_cpl}
\end{figure*}

\begin{algorithm*}[t]
\small
\begin{multicols}{2}
\begin{algorithmic}[1]
\Function{$(\bf F_O, \bf f_O, \bf suidx)$ = CPL}{{$\bf F_S$}, {$\bf F_I$}, $k$}

\For{$i = 0$ \textbf{to} ncols($\bf F_S)-1$} \Comment max pooling
		\State ${\bf f_{max}}[i]$ = max$({\bf F_S}[:,i])$ \label{op_cpl_max}
        \State ${\bf idx}[i]$ = argmax$({\bf F_S}[:,i])$
\label{op_cpl_argmax}
\EndFor
      \State ${\bf uidx}$ = unique(${\bf idx}$)  \Comment  set operation \label{op_cpl_set}
      \State $\bf f_S[j]$ = $\sum_{\bf idx[i]=uidx[j]} {\bf f_{max}[i]}$ \label{op_cpl_fset}
      
      \If {WCPL}
      	\State $\bf fr[j]$ = $\left\vert\{i \:|\: {\bf idx}[i]={\bf uidx}[j]\}\right\vert$  \Comment frequency of ${\bf uidx}[j]$ in ${\bf idx}$
      \EndIf
      	
      \State ${\bf -, l}$ = sort(${\bf f_S}$) \Comment sorting
      \State ${\bf suidx}$ = ${\bf uidx[l]}$ \label{op_cpl_sortind}
      
      \If {WCPL}
      \State${\bf midx}$ =  repeat$({\bf suidx, fr})$
      \State ${\bf rmidx}$= resize(${\bf midx}$, $k$) \Comment nearest-neighbor resizing
      \State $\bf F_O = {\bf F_I}[\bf rmidx,:]$ \Comment point collection
      \Else
      \State ${\bf rsuidx}$= resize(${\bf suidx}$, $k$) \Comment nearest-neighbor resizing \label{op_cpl_resize}
      \State $\bf F_O = {\bf F_i}[\bf rsuidx,:]$ \Comment point collection
      \EndIf
      
\For{$i = 0$ {\textbf{to}} ncols(${\bf F_O})-1$} \Comment max pooling
        \State ${\bf f_O}[i]$ = max(${\bf F_O}[:,i]$)
\EndFor \\
\If {WCPL}

\Return $(\bf F_O, \bf f_O, \bf rmidx)$

\Else

\Return $(\bf F_O, \bf f_O,  \bf rsuidx)$

\EndIf
\EndFunction

\end{algorithmic}
\end{multicols}
\caption{\label{alg_cpl} (Weighted) Critical Points Layer (CPL/WCPL)}
\end{algorithm*}

The block diagram of the proposed Critical Points Layer (CPL) is illustrated in Figure \ref{fig_cpl}. 
In order to elaborate more on the functionality of CPL, its pseudo-code is also provided in Algorithm~\ref{alg_cpl}. The steps of the algorithm are explained in more details as follows:

\begin{enumerate}
	\item The input point cloud $\bf{F}_S$ is a matrix with $n$ rows (corresponding to $n$ input points) and $d$ columns (corresponding to $d$-dimensional feature vectors).
	
	\item In the first step (Operation~\ref{op_cpl_max}), the maximum feature value is obtained for each column of the matrix $\textbf{F}_S$.  This is the same as the max-pooling operation in PointNet \cite{qi2017pointnet}. The resulting $d$-dimensional feature vector, denoted by ${\bf f_{max}}$, has the same dimension as input feature vectors and can be independently used for classification and segmentation tasks. However, we are interested in down-sampling the input points rather than generating a single feature vector out of them. 
	To this aim, the index of each row with a maximum feature value is also saved in the index vector $\bf{idx}$. Vector $\bf{idx}$ contains the indices of all the points that have contributed to the feature vector ${\bf f_{max}}$. 
%- Due to sucess of max pooling layers in PointNet and so on. 
By definition, we call these points, the \textit{Critical Points} (CP). These are the important points that should be preserved in the down-sampling process.  

	\item Index vector $\bf idx$ may contain multiple instances of the same point. To avoid these repetitions, unique indices are extracted from $\bf idx$, using the ``set (unique)'' function (Operation~\ref{op_cpl_set}). Output set which has the unique indices is called the \textit{Critical Set} (CS) and is denoted by $\bf uidx$. Beside finding the unique vector, we also add-up the feature values from ${\bf f_{max}}$ that correspond to the same point or index (Operation~\ref{op_cpl_fset}). Resulting feature vector $\bf f_S$ will be later used to sort the input points.

	\item Next, feature vector $\bf f_S$ is sorted (in an ascending order). Corresponding indices in $\bf uidx$ are also rearranged based on the sorting output (Operation~\ref{op_cpl_sortind}), resulting in an index vector which is denoted by $\bf suidx$.	This step is necessary for the following sampling (resizing) operation. It also makes CPL invariant to the order of input points.

	\item Number of elements in $\bf suidx$ may  differ for different point clouds in the input batch. For batch processing however, these numbers need to be the same. To address this, for each point cloud in the input batch, the index vector $\bf suidx$ is up-sampled to a fixed size vector $\bf rsuidx$ using an  up-sampling method for integer arrays, such as \textit{nearest neighbor} resizing (Operation~\ref{op_cpl_resize}).   

	\item As the final step, the up-sampled index vector $\bf rsuidx$, which contains the indices of all the critical points, is used to gather points and their corresponding feature vectors. Since different feature vectors may correspond to a single point, and because of the information being filtered in hidden NN layers, we may want to gather the features from other layers (denoted by $\bf F_I$) than those used for selecting the points (denoted by $\bf F_S$). However, critical points are defined based on the contribution of each point in the maximum feature vector obtained from $\bf F_S$, thus here we use ${\bf F_I}={\bf F_S}$.
\end{enumerate}

One of the main requirements of any layer designed for point cloud processing is its invariance to point cloud permutation. The proposed CPL, fulfills this requirement via following properties:
\begin{itemize}
 \item Sorting the feature vector $\bf f_S$ in step $4$ is order independent, because sorting is based on feature values and not based on indices of the points.
 \item Nearest-neighbor resizing in step $5$ is invariant to swapping the index of the input points, i.e.
 \begin{equation}
 	resize(sort(swap({\bf uidx})))=
 	swap(resize(sort(\bf uidx)))
 \end{equation}
 where $sort$ is applied based on feature values, and $swap$ is applied on index only. 
\end{itemize} 

\begin{figure*}%[h]
\centering
\includegraphics[trim = 0in 0in 0in 0in, clip, width=4.5in]{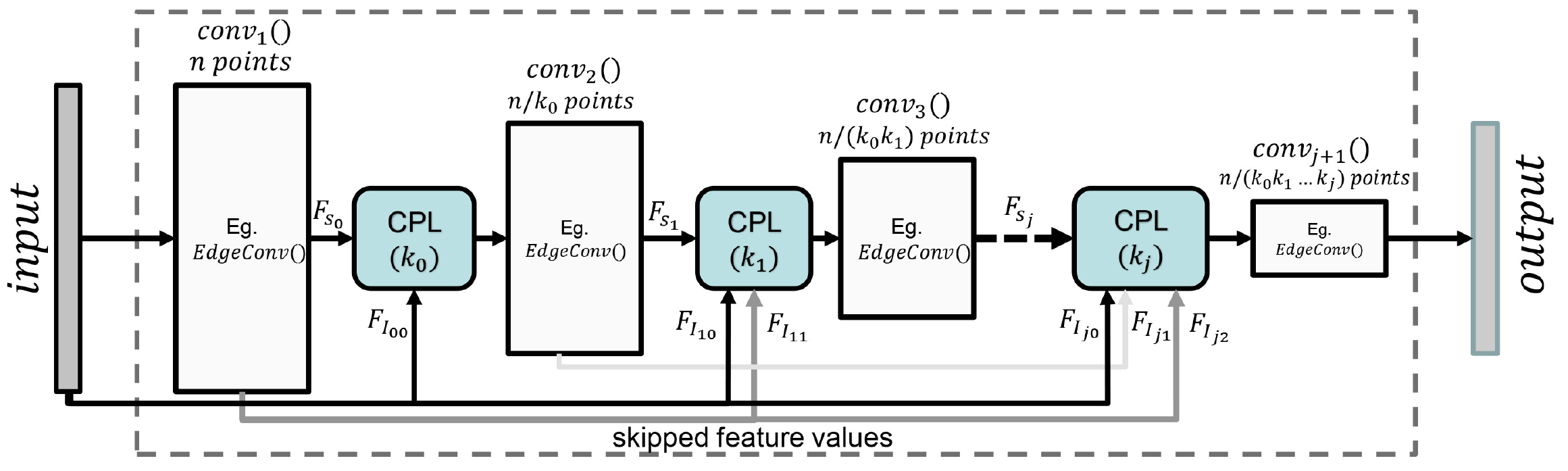}
\caption[WCPL]{General block diagram of the proposed CP-Net. }
\label{fig_cpnet}
\end{figure*}

\subsection{Weighted Critical Points Layer (WCPL)}
In CPL, a point in the point cloud is counted as a critical point if any of its features contributes to the output maximum feature vector ${\bf f_{max}}$, regardless of the number of its contributing features. For example, if a point contributes with two of its features, while another point has ten contributing features, both are treated the same in CPL. In other words, in CPL the ``\textit{importance}" of a point has a binary value: a given point is either important (critical) or unimportant (uncritical). In this section, we introduce a modified version of CPL, called Weighted Critical Points Layer (WCPL). The proposed WCPL assigns weights to points based on their level of contribution to ${\bf f_{max}}$. 

In this context, to increase the weight of a point by a factor of $C$, we repeat the point index $C$ times. By increasing the repetition frequency, the probability of selecting the point in the down-sampling process will also increase. From another point of view, in WCPL, the probability of missing a critical point in the output is lower than that in CPL. The pseudo-code of WCPL is given in Algorithm~\ref{alg_cpl} using the \textbf{if} statements.

%\begin{figure}[h]
%\centering
%\includegraphics[trim = 0in 0in 0in 0in, clip, width=3.4in]{figs/alg_wcpl.jpg}
%\caption[WCPL]{WCPL}
%\label{fig_alg_wcpl}
%\end{figure}

\begin{figure*}%[t]
\centering
\includegraphics[trim = 0in 0in 0in 0in, clip, width=5.0in]{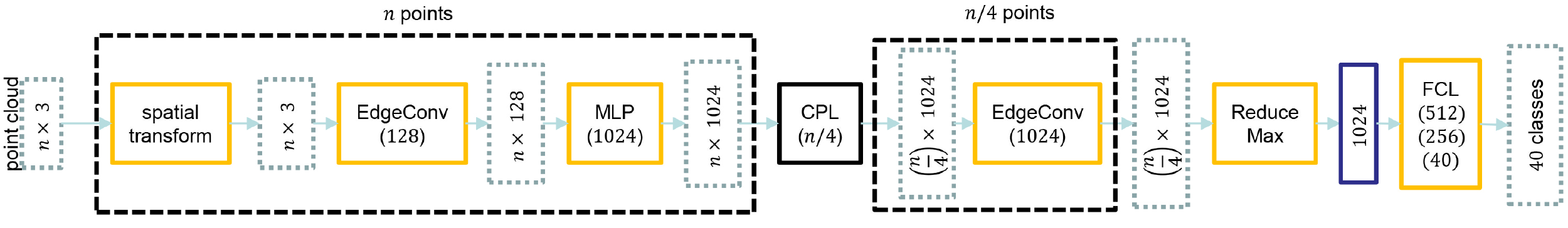}
\caption[WCPL]{The proposed point cloud classification network using CP-Net.}
\label{fig_class_cpnet}
\end{figure*}

\subsection{Critical Points Net (CP-Net)}
In this section, we propose a hierarchical architecture to apply deep convolutional neural networks to point clouds, by systematically reducing the number of points using the proposed CPL/WCPL. In the proposed network model, named \textit{Critical Points Net} (CP-Net), any graph convolution method, such as DCNN \cite{atwood15}, GCNN \cite{masci2015geodesic}, MoNet \cite{monti2017geometric} or EdgeConv (from DGCNN \cite{wang2018dynamic}) can be used in convolution layers. The block diagram of CP-Net using EdgeConv as an example is shown in Figure~\ref{fig_cpnet}.

The input in Figure \ref{fig_cpnet} is an unordered point cloud of size $n$. In the first step, the point cloud is passed into a convolution layer of choice to filter the input into a richer set of features.
The filtered output point cloud $F_{S_0}$ is then used as an input to the first CPL/WCPL. Using a CPL/WCPL with down-sampling factor $k_0$, the number of points in $F_{S_0}$ is reduced to $n/k_0$ points. These steps are repeated for as many times as necessary to achieve a desired size of the point cloud (both in terms of number of points and feature vector size). Note that at $j$-th CPL/WCPL block, one can also benefit from using or concatenating features from all or some of the previous layers, i.e., $\{F_{I_{j0}},F_{I_{j1}},…,F_{I_{jj}}\}$, as long as they correspond to the same points. As a result, the number of output points will be $\frac{n}{k_0 k_1…k_{j+1}}$.

\subsection{CP-Net for 3D Object Classification}
\label{cpnet_classification}
Here we give an example of CP-Net application in the $3$D object classification problem. Block diagram of the proposed network is illustrated in Figure~\ref{fig_class_cpnet}. The network is composed of three subnets: 1) $n$-point feature extraction subnet, 2) $(n/4)$-point subnet and 3) classification subnet. The detailed steps of the proposed network are as follows:
\begin{itemize} %[label=(\alph*)]
	\item Network input is an unordered point cloud of size $n\times 3$, where each point is a $3$D vector.
	
	\item The input data goes through a spatial transformer network as explained in~\cite{qi2017pointnet}, to make it robust against any rigid transformation, including rotation and translation. It is worth noting that instead of using the original input, a modified version of EdgeConv~\cite{wang2018dynamic} edge feature is used for spatial transformation, as explained in the next step \footnote{The proposed layer can be efficiently used along with other complex graph convolution layers such as EdgeConv \cite{wang2018dynamic} or MoNet  \cite{monti2017geometric}. In this paper, we use a modified version of EdgeConv~\cite{wang2018dynamic}, as a graph convolution operator.}.
	
	\item The output of the spatial transform goes into a filtering CNN, here EdgeConv \cite{wang2018dynamic}, to produce richer features. Unlike the original EdgeConv \cite{wang2018dynamic} operator which uses two kernels in the edge feature function, we use the triple-kernel version $h_{\Theta}(x_i,x_j-x_i,(x_j-x_i)^2)$, where $(x_j-x_i)^2$ is element-wise square operation between each point $x_i$ and its neighbouring point $x_j$. In the proposed network, applying the EdgeConv with $128$ filters to the input point cloud of size $n\times 3$, results in a point cloud of size $n\times 128$.
	
	\item  A multi-layer perceptron (MLP) layer expands the feature dimension from $128$ to $1024$ features, resulting in a point cloud of size $n\times 1024$. 
	
	\item Next, CPL/WCPL is applied to find the critical points and to reduce the number of input points. As shown in Section~\ref{sec_exp}, this step reduces the computational complexity without any loss in the classification accuracy. A down-sampling factor of $1/4$ is chosen to reduce the number of points from $n$ to $n/4$. 

	\item Another EdgeConv layer is used to filter the point cloud, this time by preserving the depth and size to further process the received point cloud. Note that reducing the number of points in the previous layer highly reduces the computational complexity of the this layer.

	\item A reduce-max layer is used to generate a vector of size $1024$, out of the point cloud of size $n\times 1024$.
	
	\item Finally, fully connected layers of size $512, 256$ and $40$ are applied to transform the feature vector of size $1024$ to the number of classes in the ModelNet$40$ dataset \cite{wu20153d}, which is $40$. 
\end{itemize}

In the proposed $3$D classification method, standard softmax cross entropy is used as the loss function. In addition, all layers include a ReLU activation function and batch normalization.

\section{Experiments}
\label{sec_exp}
\subsection{Data preprocessing}
We evaluate our model on ModelNet$40$ $3$D object classification dataset \cite{wu20153d}. The dataset contains $12,311$ meshed CAD models from $40$ different object categories out of which $9,843$ models are used for training and $2,468$ models for testing. From each model mesh surface, $1024$ points are uniformly sampled and normalized to the unit sphere. For data augmentation, we randomly scale, rotate and shift each object point cloud in the $3$D space. 

\subsection{Training Details}
To train the model, we use Adam optimizer with an initial learning rate $0.001$ and exponentially decay it with a rate of $0.5$ every $200,000$ steps. The decay rate of batch normalization starts from $0.5$ and is increased to $0.99$. The Dropout with probability $0.5$ is used in the last two fully-connected layers. Training the network with TensorFlow on an Nvidia P$100$ GPU with batch size $32$, takes $9-10$ hours for $400$ epochs.

\subsection{Statistical Results}
To evaluate the performance of a $3$D point cloud classification method, we use both \textit{overall accuracy} and \textit{per-class average accuracy}, calculated over all the test samples. 

The classification accuracy results for our proposed CP-Net/WCP-Net are shown in Table~\ref{tab_class_cpnet_results} with comparisons against the previously proposed methods. As illustrated, our CP-Net/WCP-Net methods rank as the runner-up to \cite{liu2019relation} and surpass the accuracy of all other methods in Table~\ref{tab_class_cpnet_results} and of ModelNet$40$ benchmark leader board.

\begin{table}%[h]
%\tiny
\centering
\resizebox{\columnwidth}{!}{
\begin{tabular}{lcc}
\hline
Algorithm & Overall Accuracy (\%)  & Mean Class Accuracy (\%) \\ \hline
Vox-Net \cite{maturana2015voxnet}    & $83.00$  &    $85.9$ \\
ECC \cite{simonovsky2017dynamic}                         & $83.2$ & - \\
SO-Net \cite{li2018so}                       & $89.16$  & -  \\
Pointnet \cite{qi2017pointnet}                     & $89.20$  &    $86.0$    \\ 
Pointnet++ \cite{qi2017pointnet++}                   & $90.70$  &   -     \\
KCNet \cite{shen2018mining}                        & $91.0$    & -    \\
KD-Net \cite{klokov2017escape}                      & $91.8$ & - \\
DGCNN\cite{zhang2018end} (1 vote)             & $91.84$   &   $89.40$   \\ 
RS-CNN \cite{liu2019relation}                 & $93.6$ & - \\
 
\hline
%Ours (BASELINE)& $90.50$   & $87.30$  \\ 
Ours (CP-Net)                        & $92.33$    &   $89.90$  \\ 
Ours (WCP-Net)                       & $92.41$   &  $90.53$    \\ 
\hline
\end{tabular}
}
\caption[CPNET]{Classification accuracy results on ModelNet40 dataset \cite{wu20153d}, for input size $1024\times3$.}
\label{tab_class_cpnet_results}
\end{table}

%\begin{enumerate}[label=(\alph*)]
%\item Performance on smaller classes?

%\item Number of votes to get the results

%\item Compare some of the results for different classes, between pointnet, DGCNN and cp-net
%\end{enumerate}

\begin{figure*}[!t]
\centering
\includegraphics[trim = 0in 0in 0in 0.0in, clip, width=4.5in]{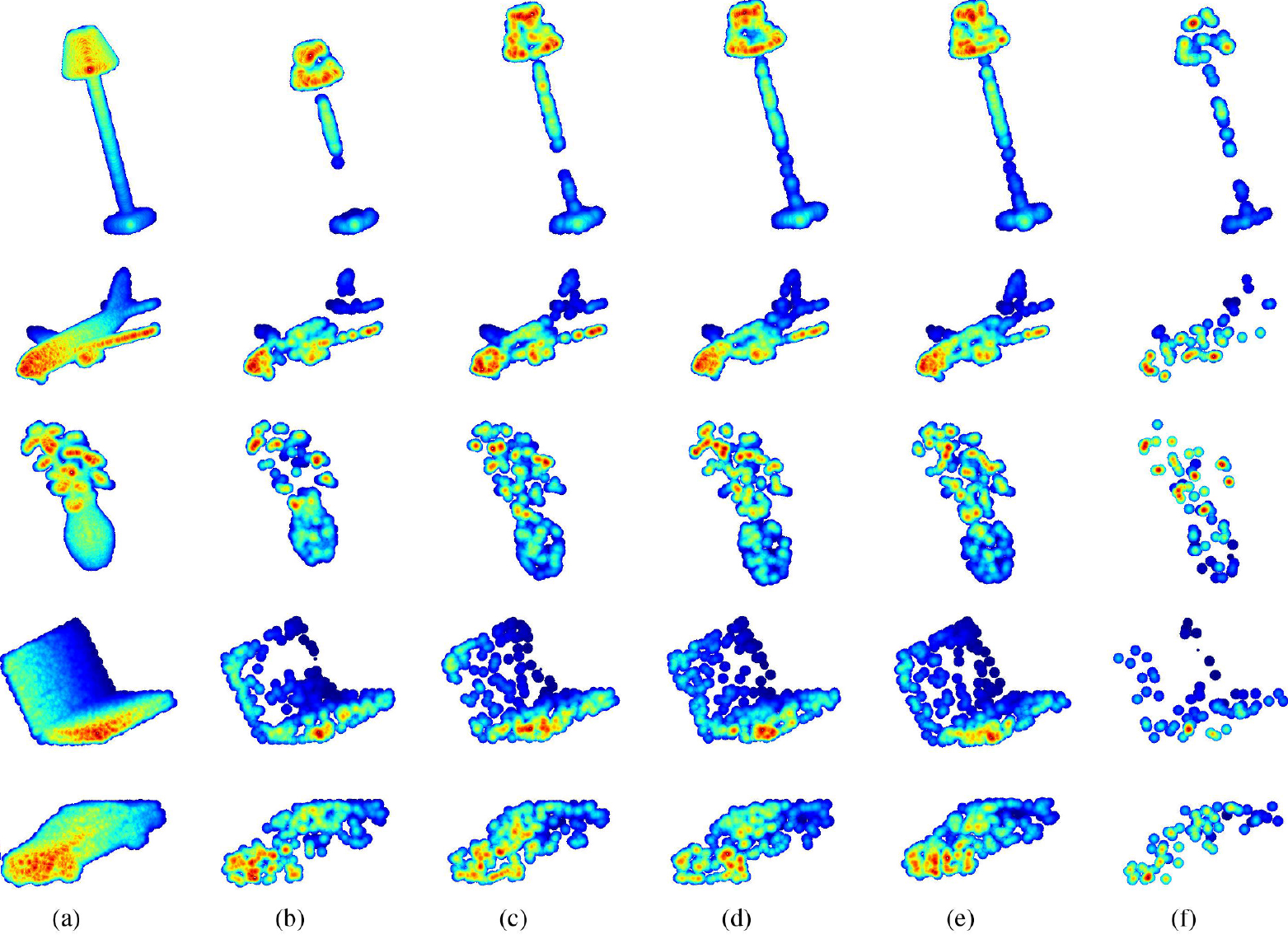}
\caption{(From top to bottom and left to right) The original point clouds for the laptop and car object categories in ModelNet$40$ dataset~\cite{wu20153d}, and their down-sampled versions (with ratio $\frac{1}{4}$) obtained after training the classification CP-Net, shown in Figure~\ref{fig_class_cpnet}, for $1$, $100$, $200$ and $300$ epochs. (f) The result of training for $300$ epochs with down--sampling ratio $\frac{1}{16}$. The images are color coded to reflect the depth information.}
\label{fig_cpl_vis_examples}
\end{figure*}

\subsection{Qualitative Results}
Figure~\ref{fig_cpl_vis_examples} shows how the proposed CPL learns to down-sample different point clouds. The original point clouds of size $1024$ for the object classes \textit{lamp}, \textit{airplane}, \textit{flower-pot}, \textit{laptop} and \textit{car} are shown in Figure~\ref{fig_cpl_vis_examples}(a). Figures~\ref{fig_cpl_vis_examples}(b-e) correspond to the outputs obtained using the down-sampling ratio of $0.25$, at epochs $1$, $100$, $200$ and $300$, respectively. As seen in Figure~\ref{fig_cpl_vis_examples}(b), at the beginning of the training, some important parts of the objects, such as lamp column, flower leaves and laptop screen are partially lost as a result of down-sampling. After $300$ epochs of training however, CPL learns to down-sample the point cloud such that the \textit{critical points} of the object are mostly retained. In the context of point cloud  classification, by \textit{important points} of an object we mean those points that contain the necessary information to discriminate between different objects in the dataset.

Figure~\ref{fig_cpl_vis_examples}(f) shows the corresponding point clouds down-sampled by the ratio of $\frac{1}{16}$, after $300$ training epochs. As seen, the important points of each object for our classification task are still preserved even in such small $64$-point point clouds. The lamp column, airplane wings and six corners of the laptop are some examples of the preserved important object parts. 

\subsection{Ablation studies}
{\bf EdgeConv Kernels}
The effect of using two and three kernels (in EdgeConv operator used in DGCNN~\cite{wang2018dynamic}) on the overall classification accuracy and execution time is shown in Table~\ref{tab_kernels}. For double-kernel version, we use the one used in ~\cite{wang2018dynamic}. The triple-kernel version is defined in section~\ref{cpnet_classification}-c. As seen, the triple kernel version is computationally more complex than the double kernel version. In both cases, not only the proposed  CP-Net/WCP-Net outperforms the DGCNN in classification accuracy, it is computationally less complex, due to CPL/WCPL point cloud down-sampling.

\begin{table}%[h]
\small
\centering
%\resizebox{\columnwidth}{!}{
\begin{tabular}{lcc}
\hline
Method  & Double Kernel& Triple Kernel \\ \hline
DGCNN & 91.84 (135ms) & 89.26 (141ms)\\
CP-Net &  \textbf{91.88} (115ms) & 92.33 (119ms)\\
WCP-Net & 91.76 (116ms) & \textbf{92.41} (120ms)\\
\hline
\end{tabular}
%}
\caption{Effect of edge feature kernels on overall classification accuracy (\%) and execution time (in ms).}
\label{tab_kernels}
\end{table}

{\bf Effect of Bottleneck Dimension}
The effect of bottleneck layer size (number of features in the output feature vector) on classification accuracy is shown in Table~\ref{tab_bottleneck}. Clearly, increasing the  bottleneck layer size improves the accuracy, however it almost saturates at around $1024$ features. Note that even with the bottleneck size of $64$, accuracy of the proposed CP-Net ($\%89.35$) is more than that of the PointNet with bottleneck size of 1024 ($\%89.20$). 

%In average, CP-Net yields better classification accuracy than WCP-Net.

\begin{table}[h]
\small
\centering
%\resizebox{\columnwidth}{!}{
\begin{tabular}{lcccccc}
\hline
  & 64 & 128 & 256 & 512 & 1024 \\ \hline
CP-Net & 89.35 & 89.83 & 90.85 & 91.94 & 92.33 \\
WCP-Net & 89.16 & 89.71 & 90.73 & 91.54 & 92.41\\
\hline
\end{tabular}
%}
\caption{Effect of bottleneck dimension on accuracy (\%).}
\label{tab_bottleneck}
\end{table}

{\bf Effect of Down-Sampling Ratio}
Table~\ref{tab_ds_ratio} shows the effect of down-sampling ratio on classification accuracy. As expected, the more the point cloud is shrunk, the lower the accuracy is. This is because some of the important information about the object is lost as a result of down-sampling. The proposed CPL and WCLP layers however, preserve the important object points as much as possible. This can be verified from Table~\ref{tab_ds_ratio}, where the difference between the accuracy values at down-sampling ratios $1$ and $1/16$ (corresponding to point clouds of size $1024$ and $64$ points) is only $\%0.73$. This means that in the down-sampling process, CPL preserves the most important information of each object, so that with such small number of points, objects are still classified with high accuracy.

Taking the effect of down-sampling for a ratio of $4$, i.e., $256$ points, it is worth comparing the accuracy of CPL with a random down-sampler. If we use random down-sampling of ratio $4$ instead of CPL/WCPL, the accuracy of the CP-Net classification network drops to $91.47$ from $92.33$ and $92.41$ for CPL and WCPL, respectively, which shows the effectiveness of CPL in obtaining the critical points and features in the architecture. It is worth noting that although using random down-sampling is viable in training, it is not mathematically and statistically sound at inference. The reason is that every time the inference is run, the random sampler will produce a new set of indices for down-sampling, resulting in a non-deterministic output/classification for the same input.

\begin{table}[h]
\small
\centering
%\resizebox{\columnwidth}{!}{
\begin{tabular}{lccccc}
\hline
  & 1 & 1/2 & 1/4 & 1/8 & 1/16 \\ \hline
CP-Net  & 92.25 & 92.24 & 92.33 & 92.29 & 91.52 \\
WCP-Net & 92.09 & 92.15 & 92.41 & 92.03 & 91.81 \\
\hline
\end{tabular}%}
\caption{Effect of down-sampling ratio on accuracy (\%).}
\label{tab_ds_ratio}
\end{table}

{\bf Time and Space Complexity}\\
%Plot (Accuracy vs Time Complexity (Execution Time)) \\
%Plot (Accuracy vs Space Complexity) \\
%CP-Net vs when removing CPL \\
Table~\ref{tab_complexity} compares the execution time of running the proposed CP-Net/WCP-Net with other state-of-the-art object classification methods, on an Ubuntu machine with a single P$100$ GPU and an Intel $1.2$~GHz CPU. Batch size of $32$ $1024$-point point clouds is used in all the experiments. 

In terms of model size, the smallest models are generated by KCNet.  CP-Net and WCP-Net generate the largest models due to their larger number of network parameters. The model size can be reduced by decreasing the bottleneck dimension.

In terms of computational complexity, CP-Net and WCP-Net run faster than both DGCNN and PointNet++. However, they are slower than PointNet and KCNet. The lower computational complexity of CP-Net in comparison with DGCNN is due to the employment of CP layer in its network. Similarly, by a proper network design, the proposed CPL/WCPL is expected to accelerate other classification deep networks, such as KCNet and PointNet++. 

\begin{table}%[h]
\small
\centering
%\resizebox{\columnwidth}{!}{
\begin{tabular}{lccc}
\hline
Method   &  Model Size & Inference Time & Accuracy \\ 
   &  (MB) & (ms) & (\%) \\  \hline
PointNet & 40 & 36.1 & 89.20\\
PointNet++ & 12 & 232.2 & 90.70\\
KCNet & \textbf{11} & \textbf{26.4} & 91.0\\
DGCNN & 21 & 134.5 &  91.84 \\
%CP-Net (BASELINE) & 24 & 55.8  & 90.50 \\
CP-Net/WCP-Net & 57 & 118.9 & \textbf{92.33/92.41}\\
\hline
\end{tabular}
%}
\caption{Model size (in MB), inference mode execution time  on P100 GPU (in ms) and classification accuracy ($\%$).}
\label{tab_complexity}
\end{table}

\section{Conclusion}
In this paper we propose a new deterministic adaptive down-sampling method that can be trained to pass on the most important points (critical points), to the next layers in a neural network. Two down-sampling layers, the critical points layer (CPL) and its weighted version, the weighted CPL (WCPL) are proposed. As a systemic approach of using CPL/WCPL in deep neural networks, CP-Net, a hierarchical implementation of the these layers, is also introduced. Finally, a deep classification network is designed based on the proposed CP-Net for $3$D object classification. Experimental results using a common dataset show the competitiveness of the proposed method in terms of classification accuracy in comparison to previous state-of-the-art methods.

CPL is an adaptive layer that can dynamically down-sample the unordered data. This property may inspire design of a new type of auto-encoders that can handle unordered data such as point clouds. This approach is already under investigation and results will be shown in a future work. 

{\small
\bibliographystyle{ieee_fullname}
\bibliography{egbib}
}

\end{document}